# MSGDD-cGAN: Multi-Scale Gradients Dual Discriminator Conditional Generative Adversarial Network

**Mohammadreza Naderi[1], Zahra Nabizadeh[2],**

**Nader Karimi[3], Shahram Shirani[4], Shadrokh Samavi[5]**

Department of Electrical and Computer Engineering, Isfahan University of Technology, Isfahan, Iran[1,2,3,5]
Department of Electrical and Computer Engineering, McMaster University, Hamilton, Canada[4,5]
mr.naderi@ec.iut.ac.ir[1], z.nabizadeh@ec.iut.ac.ir[2], nader.karimi@cc.iut.ac.ir[3], shirani@mcmster.ca[4], samavi@mcmaster.ca[5]

## Abstract

Conditional Generative Adversarial Networks (cGANs) have been used in many image processing tasks. However, they still have serious problems maintaining the balance between conditioning the output on the input and creating the output with the desired distribution based on the corresponding ground truth. The traditional cGANs, similar to most conventional GANs, suffer from vanishing gradients, which backpropagate from the discriminator to the generator. Moreover, the traditional cGANs are sensitive to architectural changes due to previously mentioned gradient problems. Therefore, balancing the architecture of the cGANs is almost impossible. Recently MSG-GAN has been proposed to stabilize the performance of the GANs by applying multiple connections between the generator and discriminator. In this work, we propose a method called MSGDD-cGAN, which first stabilizes the performance of the cGANs using multi-connections gradients flow. Secondly, the proposed network architecture balances the correlation of the output to input and the fitness of the output on the target distribution. This balance is generated by using the proposed dual discrimination procedure. We tested our model by segmentation of fetal ultrasound images. Our model shows a 3.18% increase in the F1 score comparing to the pix2pix version of cGANs.

## Introduction

Image synthesis consists of generating an image in a specific domain conditioned or unconditioned to input information. One of the most influential models for synthesizing or translating images is Generative Adversarial Networks (GANs) (Goodfellow et al. 2014) which has received plenty of attention since their first introduction in 2014. The models presented for image synthesizing could be applied to various image processing tasks; however, to achieve maximum performance in each image processing category, the image synthesizing models need to become task-specific. For instance, in (Zhang et al. 2018), authors fit a pre-trained segmentation network to GANs framework to achieve better semantic segmentation results. References (Kumar, Bhandarkar, and Prasad 2018) and (Gwn Lore et al. 2018) using the adversarial training procedure to apply their specific models to the depth estimation task. Authors of (Zhu et al. 2018) also use this framework in the image classification task.

Numerous works change the architecture and hyper-parameters of GANs. Works of (Zhang et al. 2019; Choi et al. 2018; Zhu et al. 2017; Yi et al. 2017; Karras, Laine, and Aila 2019; Karnewar and Wang 2020) concentrate on changing and adjusting the architecture of GANs. Also, the authors of (Mao et al. 2017) and (Arjovsky, Chintala, and Bottou 2017) have worked on the training procedure of GANs. Recently Karnewar and Wang proposed MSG-GAN (Karnewar and Wang 2020), which solves the generator collapse problem of the GAN model, as mentioned in (Salimans et al. 2016; Wang et al. 2017; Yazıcı et al. 2019). Karnewar and Wang's solution is by boosting the flow of gradients from the discriminator to the generator. They used multiple connections between generator and discriminator architectures. These connections also help reduce vanishing gradients problems in GAN's architecture, leading to a stable model training procedure and performance. MSG-GAN (Karnewar and Wang 2020) learns image synthesizing from input noise. We can replace the GAN with a conditional GAN in the MSG-GAN framework. We also will show that connections between the generator and the discriminator could help conditional GAN work more stable during training. This stability will allow us to balance the network by using dual discriminators.

Using multiple connections between the generator and discriminator in MSG-GAN, allows the gradients to flow better from the discriminator to the generator. However, in cGANs, the model must construct the output conditioned on input information. Therefore if we use multiple connections and gradient boosted models, they may be lead to overfitting to the target domain images and lose most of the correlation to the input. In other words, the output is constructed from noise-like feature vectors, which results from paying the network's full attention to output losses. On the other hand, if we condition the output during the training time on the input too much, the model will fail during the test time. The network cannot generalize the learned information in the test time because it has learned a high correlation between the input and output. Hence, maintaining the balance between passing information from the input to the output of the network and the intensity of the gradient backpropagation is a challenging and important task. This paper proposes a refined architecture for the conditional GAN (Mirza and Osindero 2014) category. We call the proposed model multi-scale gradient dual discriminators conditional GAN (MSGDD-cGAN). It resolves explained problems by utilizing as much helpful information as possible from the input and conditioning the multi-connections gradient boosted model on that information. We will also show that our method achieves better results during the test time due to saving this balance.

The rest of this paper is organized as follows: In the proposed platform section, we explain our generator and discriminators architectures. Then, in the experimental results section, we explain our implementation detail and experiments and explain obtained results. Finally, in the conclusion section, we discussed more the cause of the obtained results.

# Proposed Platform

We conduct our experiments on the MSGDD-cGAN framework consisting of a U-Net (Ronneberger, Fischer, and Brox 2015) generator architecture. Figures 1 and 2 shows an overview of our MSGDD-cGAN architecture, which we will explain in more detail in this section.

## Generator Architecture

Let divide the generator (G) into two general blocks, E (encoder) and D (decoder), as shown in Fig. 1. The E is constructed base on different convolution blocks, which each one is named as $CB_i$. We derive multiple outputs from different layers of E (shown in Fig. 1 as the $EO_i$) by sending the input image to the E block and using one output layer ($EOL_i$) in each stage. Each output could be modeled as Equation 1.

$$EO_i = EOL_i(CB_{i+1}(CB_i(\ldots(CB_1(Input))))) \qquad (1)$$

where $EO_i$ shows the $ith$ output layer of the E, $CB_i$ is a convolutional block, input shows the input of the generator and $EOL_i$ shows the $ith$ output layer of the encoder. These variables are shown in Fig. 1.

The outputs of D ($DO_i$) are constructed similar to $EO_i$ except that we are using upsampling instead of striding. We name these blocks as $UCB_i$, and multiple outputs ($DO_i$) are derived by sending the latent code (middle feature vectors of the G called $Z$) to D and different stage output layers. Each output could be modeled, as shown in Equation 2.

$$DO_i = DOL_{i+1}(UCB_{i+1}(UCB_{i+2}(\ldots(UCB_4(Z))))) \qquad (2)$$

where $DO_i$ is the $ith$ output of the D, $UCB_i$ is an upsampling convolutional block, $Z$ shows features extracted from the middle of the generator and $DOL_i$ shows the $ith$ decoder output layer. All of these parameters are shown in Fig. 1.

## Architecture of Dual Discriminators

Two discriminators are used in the proposed model to control the flow of information from the input to the generator's output. Controlling The flow of information with two discriminators has three effects on the output: 1) the output saves as much as correlated information to the input, 2) the output fits to the distribution of target domain, 3) the output could be compared pixel by pixel to the paired ground truth while the generality of learned information by the generator is preserved. We call the discriminator of the encoder, Dis-E, and the discriminator of the decoder, Dis-D. These discriminators are shown in Fig.1. The architecture of discriminators is inspired by the pix2pix model (Isola et al. 2017), with some modifications.

### Discriminator of Decoder (Dis-D)

Because we want to handle multiple inputs, we increase the capacity of the discriminator by doubling the number of convolution layers in each of the downsampling blocks. In fake mode, the output is passed to the discriminator as the first input. The decoder's outputs ($DO_i$) in each layer are concatenated with feature vectors in each convolution block. In real mode, ground truth is passed to the discriminator as the first input. Next, the downsamples of the ground truth in each layer are concatenated with feature vectors in each convolution block. The concatenated vector is then passed to the remaining convolution blocks of Dis-D. As discussed in (Karnewar and Wang 2020), this helps gradients flow from Dis-D to the generator from multiple paths. Thus, a better flow of gradients makes the generator fit better to the target distribution and stabilizes the generator's performance.

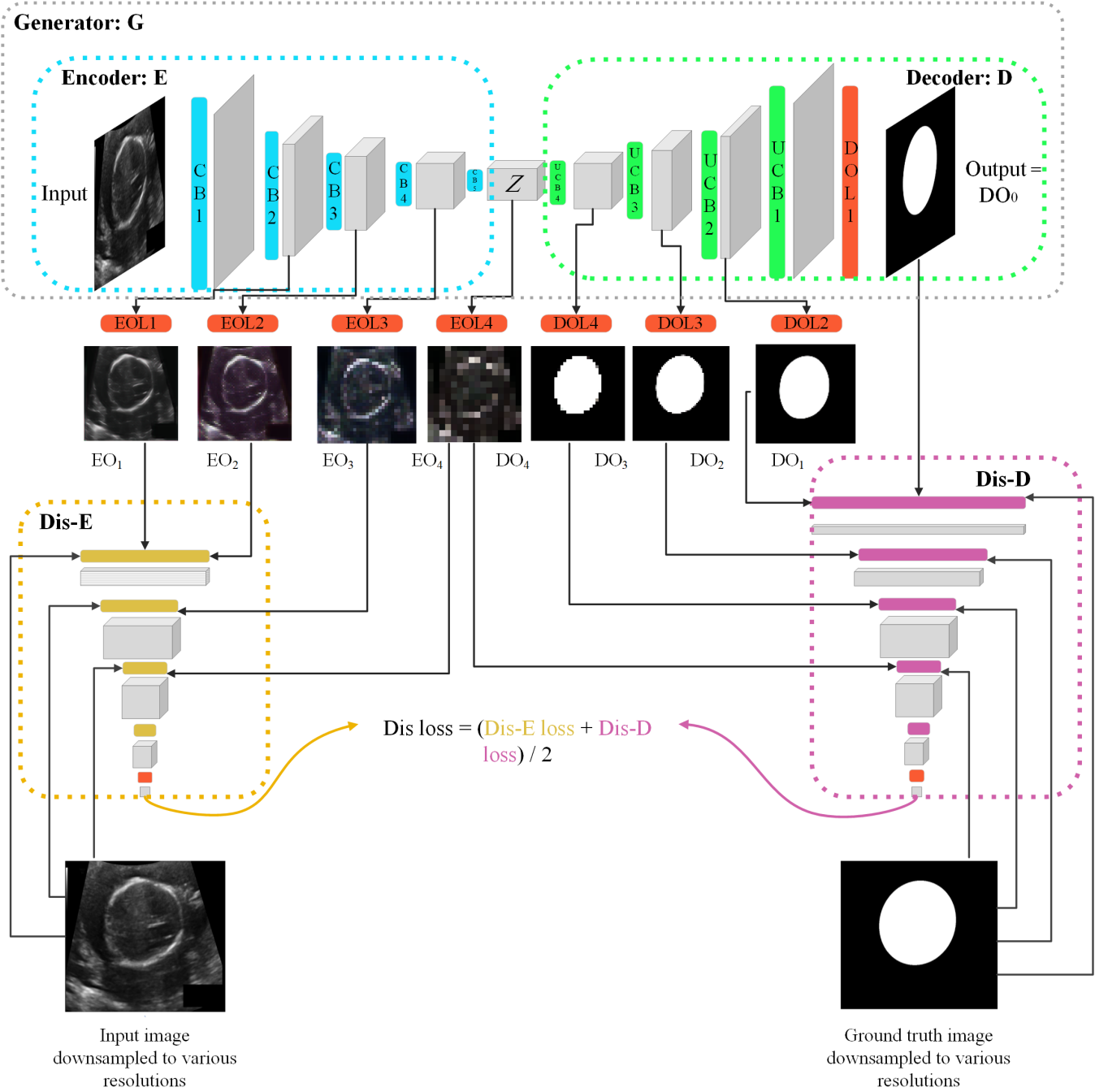

Figure 1: Architecture of MSGDD-cGAN for segmentation of fetal head images. Our architecture includes a generator (dashed gray box) and two discriminators. The generator is divided into two parts, encoder (dashed blue box) and decoder (dashed green box), for better explanations. Discriminator of encoder (Dis-E) and discriminator of decoder (Dis-D) are also shown in the dashed yellow, and dashed purple boxes respectively.

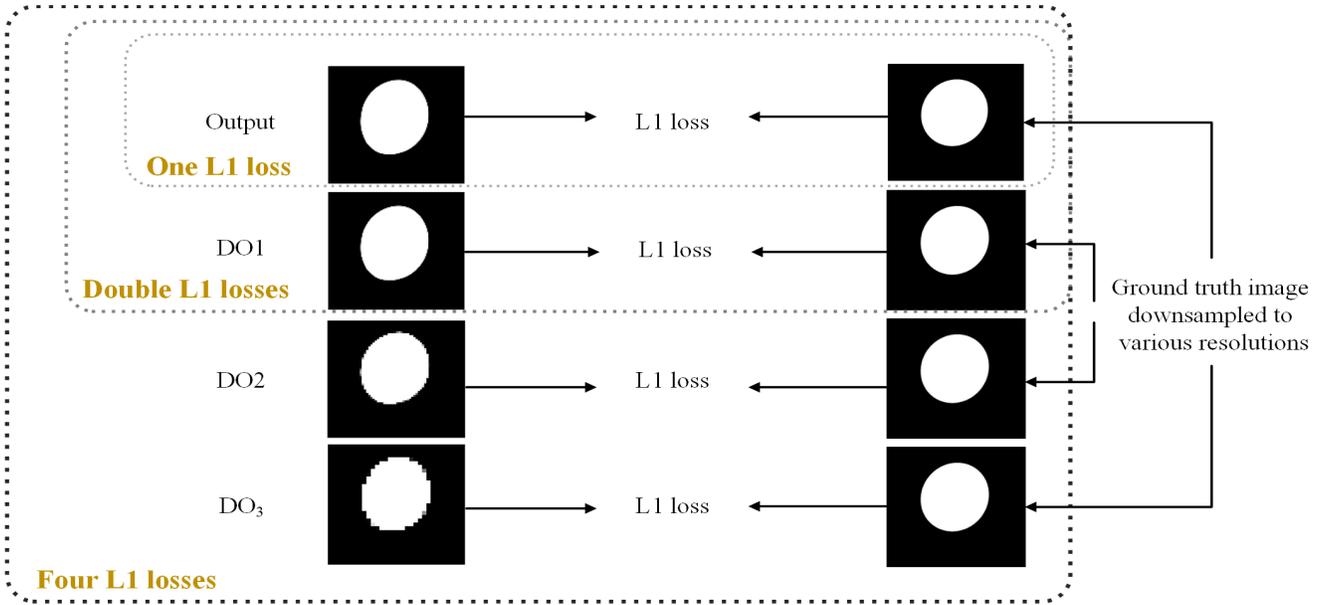

Figure 2: Multiple L1 losses computed between downsamples of ground truth and decoder's outputs.

**Discriminator of Encoder (Dis-E)**

We use Dis-E to oblige the generator to save as much information as possible from the input and conditioning the output on them. Moreover, the Dis-E prevents the generator from overfitting on the target distribution because it cannot forget most essential input features, which helps construct the output conditioned on them. The architecture of Dis-E, except in its first block, is similar to Dis-D. The first block of Dis-D is omitted because the input of the generator is always real. To train Dis-E, we pass downsamples of the input as real inputs and $EO$s as fake inputs to Dis-E.

Notice that the middle output of the generator ($EO_4$ is equal to $DO_4$) is passed to both discriminators. This will help the network to recognize similar features between the input and ground truth of the generator and use them to construct the output.

## Experimental Results

Since cGANs use real paired images in train and test processes, thus, to evaluating the quality of generated images, we used the popular image segmentation metric, F1-score, to compare our method to the baseline (pix2pix (Isola et al. 2017)).

## Implementation Details

We evaluate our method using the **HC18** dataset[1], which is a collection of fetal head ultrasound images. The dataset contains 999 images with 840×500 resolution. These images are used for biometrics measurements by segmentation of the fetal head. In other words, the pair of ultrasound images and segmentation masks are used as domains A and B. Then, the generator learns to construct segmentation masks (domain B) using ultrasound images (domain A) and masks (domain B) correlations. The training, validation, and test images are 699, 100, and 200, respectively.

We implemented our models using the Pytorch framework. All models were trained using Adam optimizer with a learning rate of 0.0002. We initialized parameters using a standard normal distribution. We also trained the pix2pix (Isola et al. 2017) model with hyperparameters to generate the best results.

We used two losses for the generation task, $\mathcal{L}_{G_{Dis}}$ and $\mathcal{L}_{G_{L1}}$. To compute $\mathcal{L}_{G_{Dis}}$ (loss of the generator base on results of discriminators) we first computed loss of Dis-E ($\mathcal{L}_{Dis_E}$) and loss of Dis-D ($\mathcal{L}_{Dis_D}$). Then we compute L2 norm loss for the generator (similar to least-square GAN (Mao et al. 2017)) based on Equations 3, 4, 5, and 6.

$$\mathcal{L}_{Dis_E} = \frac{1}{2}\mathbb{E}_{Ids \sim P_{data}}((Dis_E(Ids) - 1)^2) + \frac{1}{2}\mathbb{E}_{EOs \sim P_G}((Dis_E(EO) - 0)^2) \quad (3)$$

$Ids = [Ids_1, Ids_2, Ids_3, Ids_4]$

$EOs = [EO_1, EO_2, EO_3, EO_4]$

where $Ids_i$, $EO_i$ and $Dis_E$ are the downsample of the input image, the outputs of the encoder, and the discriminator of the encoder, respectively.

---
[1] http://doi.org/10.5281/zenodo.1322001

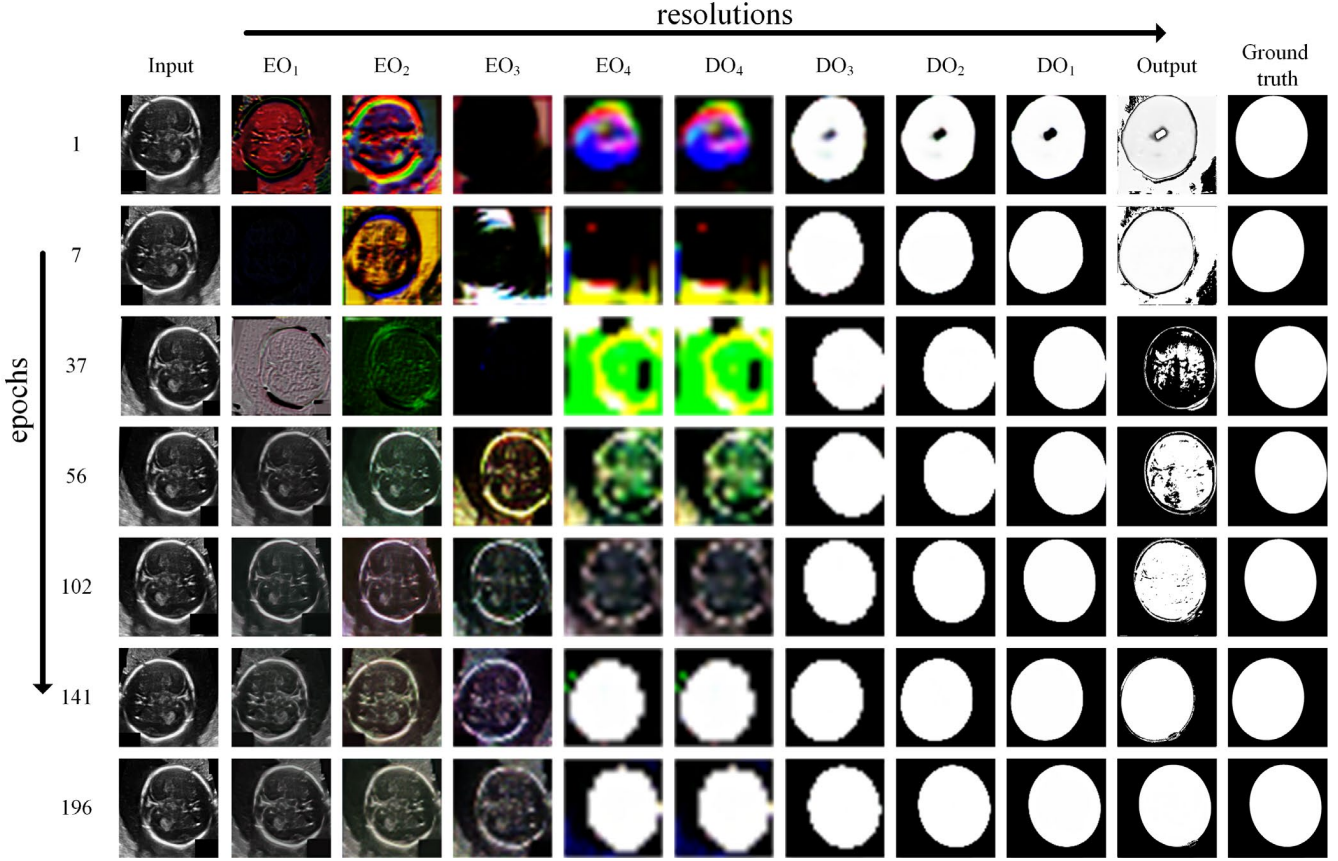

Figure 3: Training procedure for fetal head image segmentation. Each row depicts all of the network's outputs for a particular training epoch. Changes in the input and ground truth are due to image transformations in the preprocessing stage.

$$\mathcal{L}_{Dis_D} = \frac{1}{2}\mathbb{E}_{Gtds \sim P_{data}}((Dis_D(Gtds) - 1)^2) + \frac{1}{2}\mathbb{E}_{(DO) \sim P_G}((Dis_D(DO) - 0)^2) \quad (4)$$

$Gtds = [Ground\ truth, Gtds_1, Gtds_2, Gtds_3, Gtds_4]$

$DO = [generated\ output, DO_1, DO_2, DO_3, DO_4]$

where $Gtds_i$ is a downsample of the ground truth and $DO_i$ is a decoder output and $Dis_D$ is decoder's discriminator.

$$\mathcal{L}_{Dis}(G(input)) = \frac{1}{2}(\mathcal{L}_{Dis_E}(G(input)) + \mathcal{L}_{Dis_D}(G(input))) \quad (5)$$

$$\mathcal{L}_{G_{Dis}}(G, Dis_E, Dis_D) = \frac{1}{2}\mathbb{E}_{Input \sim P_{data}} ((Dis_E(G(Input) - 1))^2 + ((Dis_D(G(Input) - 1))^2 \quad (6)$$

We also have another Loss to guide the generator more, based on the L1 distance between generated outputs and real mask downsamples in domain B, as shown in Fig.2. This loss is the summation of differences between the whole of $DOs$ and downsamples of real domain B image according to Equation 7. The overall loss of the generator and discriminators also is defined in Equation 8.

$$\mathcal{L}_{G_{L1}}(G) = \mathbb{E}_{Gtds,DO}||\sum_{all\ elements}(Gtds - DO)|| \quad (7)$$

$$G^* = arg\min_G \max_{Dis_E, Dis_D}(\mathcal{L}_{G_{L1}}(G) + \mathcal{L}_{G_{Dis}}(G, Dis_E, Dis_D)) \quad (8)$$

**Results**

We compare the results of MSGDD-cGAN and pix2pix (Isola et al. 2017) using the quality of generated masks. Fig. 3 tracks $EOs$, $DOs$, and the output of an image in different epochs during training. As we can see, these connections stabilize the training procedure. Each block has a certain task, which prevents a mixture of gradients from the discriminators with that of the generator (one of the main problems of GANs (Arjovsky, Chintala, and Bottou 2017)). We also can see the balance that the network established between input information and output loss importance during the training.

**Quantitative Results**
Table 1 shows quantitative results of our method versus the pix2pix method (Isola et al. 2017) and U-Net without adversarial training. Our MSGDD-cGAN achieves a better F1 score than the respective baselines (pix2pix and U-Net). Figure 5 shows $\mathcal{L}_{G_{L1}}$ (blue) versus $\mathcal{L}_{G_{Dis}}$ (orange). As we expected, these two losses compete with each other because one of them wants to save input information while fit to output distribution ($\mathcal{L}_{G_{Dis}}$), and the other one care only about the pixel-wise output quality ($\mathcal{L}_{G_{L1}}$). Hence, the result of this competition could be considered as the balance between the input information and the effect of the output pixel-wise losses. To better show this balance, we also proposed results of single and double L1 losses on the decoder's outputs versus four L1 losses between outputs of the decoder and the downsampled versions of the ground truth, as shown in Fig. 2.
As we can see in Table 2, decreasing the number of outputs pixel-wise losses effect ($\mathcal{L}_{G_{L1}}$) while we boosted Input information preserving through the network by using Dis-E will corrupt the mentioned balance and lead the network to save redundant information of input that cannot be generalized to the test set.

**Qualitative Results**
Figure 4 shows the balance established between input and ground truth information in our architecture to generate outputs during the test time. As we can see, the network utilizes as much useful information as possible to construct the decoder's outputs. In addition, due to boosted gradients in the decoder part, the network can fit extracted features to the ground truth distribution.

Figure 6 shows generated masks using MSGDD-cGAN, pix2pix (Isola et al. 2017), and their corresponding ground truth. As we expected MSGDD-cGAN, due to using Dis-E and the effect of $\mathcal{L}_{Dis_E}$ on the generator's loss can be conditioned better on the input image. Better conditioning will result in finding more correlations between the input and output in the middle of the generator. The Dis-D, also with multi connections, will update the network parameters in the right direction due to better gradient flow from the discriminator to the generator. Multi connections also make the network more stable to the learning rate and architecture changes.

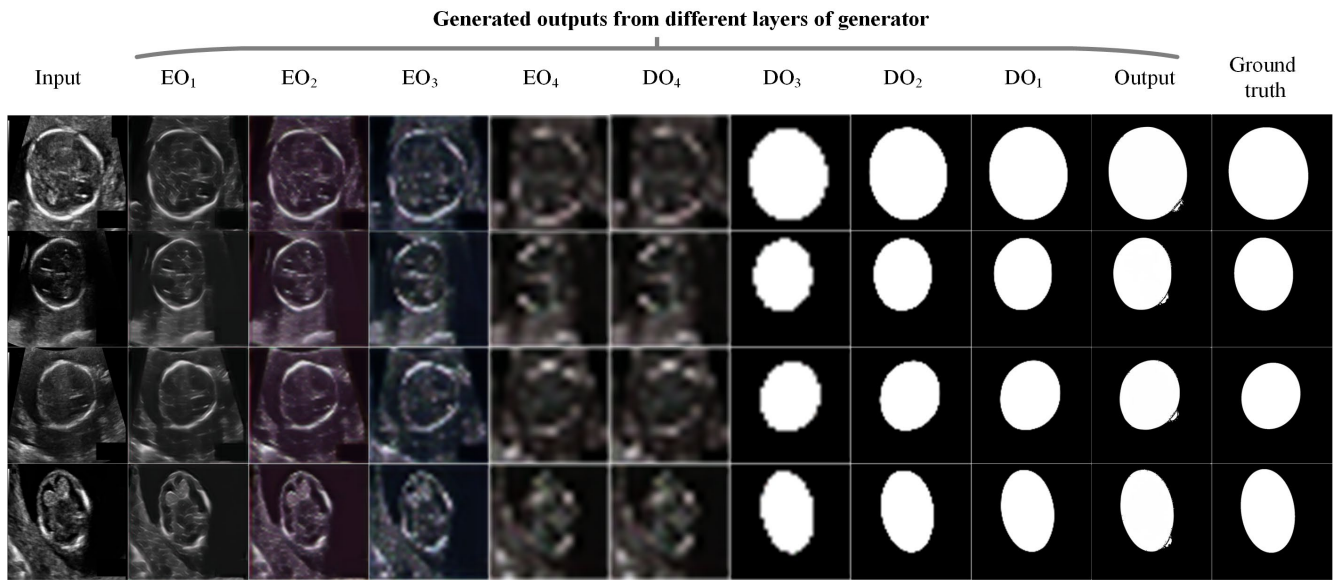

Figure 4: Results of our proposed MSGDD-cGAN technique for fetal head segmentation. The encoder and decoder generate images with different resolutions. Gradients flow directly to all levels of the encoder and decoder from the discriminators. The input has a resolution of 256×256. The resolution at each encoder output ($EO_i$) is reduced by a factor of two and increased by a factor of two at each decoder output ($DO_i$).

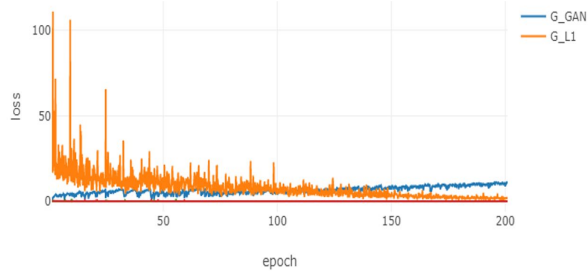

Figure 5: The convergence loss carve of our MSGDD-cGAN (blue $\mathcal{L}_{G_{Dis}}$, orange $\mathcal{L}_{G_{L1}}$)

| Models | Metric |
|---|---|
| | F1 score % |
| MSGDD-cGAN (U-Net) | **95.04** |
| pix2pix (U-Net) | 91.86 |
| U-Net without Dis | 86.06 |

Table 1: Percentage of F1-score for our method, pix2pix (Isola et al. 2017) and U-net without adversarial training (Ronneberger, Fischer, and Brox 2015).

| Models | Metric |
|---|---|
| | F1 score % |
| MSGDD-cGAN with 1L1 | 94.48 |
| MSGDD-cGAN with 2L1 | 93.19 |
| MSGDD-cGAN with 4L1 | **95.04** |

Table 2: Percentage of F1-score for our method with different L1 loss weights at the output of the architecture (shown in Fig. 2).

## Conclusion

Neural network training is challenging due to overfitting problems, learning rate adjustment, and data analysis. Overfitting to the training set means that neural networks work perfectly on the training set but cannot generalize learned information well to keep that performance on the test set. The training procedure is a major cause of overfitting. There are two ways that a cGAN model could overfit:

1) In part of the problems, neural networks find too much correlation between input and ground truth during training, which could not generalize them to the whole dataset. This will lead to worse performance on the test set versus the train set.

2) In other cases, the networks become too general and lose most of the input information to learn as much as possible of the target distribution. In other words, it cannot construct output base on enough input information.

Hence, appropriate architecture and a good definition of losses are needed to lead the training procedure in a proper direction. Then, the network can discover correlations between the input and the ground truth. In this work, we tried to define multiple losses to control the learning procedure of the generator. As a result, we could see that the network can find better general correlations between input and ground truth. Thus good results in the training time are followed by high performance in the test phase.

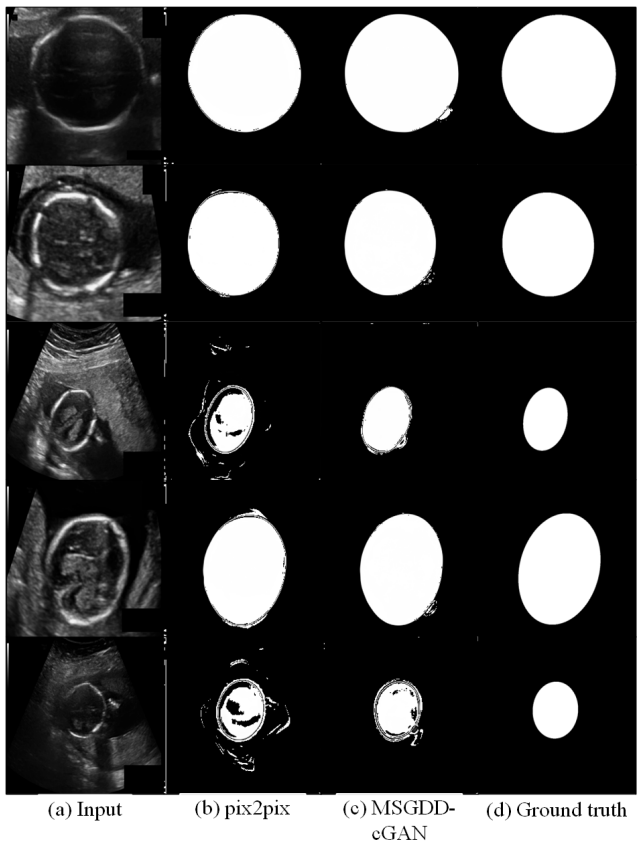

Figure 6: Comparing our method with pix2pix (Isola et al. 2017). (a) Input of network (b) Pix2pix output (c) Our method (MSGDD-cGAN) outputs (d) Ground truth